\documentclass{article}

\usepackage[square,numbers]{natbib}

     \usepackage[final]{neurips_2020_ml4ps}

\usepackage[utf8]{inputenc} 
\usepackage[T1]{fontenc}    
\usepackage{hyperref}       
\usepackage{url}            
\usepackage{booktabs}       
\usepackage{amsfonts}       
\usepackage{nicefrac}       
\usepackage{microtype}      
\usepackage{bm}
\usepackage{microtype}
\usepackage{amsmath,amssymb}
\usepackage{graphicx}
\usepackage{subfigure}
\usepackage{booktabs} 
\usepackage{multirow}
\usepackage[font=small,labelfont=bf]{caption}

\newtheorem{theorem}{Theorem}[section]

\newtheorem{lemma}[theorem]{Lemma}

\title{$\beta$-Annealed Variational Autoencoder for glitches}

\author{%
  Sivaramakrishnan Sankarapandian \\
  Proscia Inc.\\
  \texttt{siva@proscia.com} \\
   \And
  Brian Kulis \\
  Department of ECE\\
  Boston University \\
  \texttt{bkulis@bu.edu} \\
}

\begin{document}

\maketitle

\begin{abstract}
Gravitational wave detectors such as LIGO and Virgo are susceptible to various types of instrumental and environmental disturbances known as glitches which can mask and mimic gravitational waves. While there are 22 classes of non-Gaussian noise gradients currently identified, the number of classes is likely to increase as these detectors go through commissioning between observation runs. Since identification and labelling new noise gradients can be arduous and time-consuming, we propose $\beta$-Annelead VAEs to learn representations from spectograms in an unsupervised way. Using the same formulation as \cite{alemi2017fixing}, we view Bottleneck-VAEs~\cite{burgess2018understanding} through the lens of information theory and connect them to $\beta$-VAEs~\cite{higgins2017beta}. Motivated by this connection, we propose an annealing schedule for the hyperparameter $\beta$ in $\beta$-VAEs which has advantages of: 1) One fewer hyperparameter to tune, 2) Better reconstruction quality, while producing similar levels of disentanglement.
\end{abstract}



\section{Introduction}

Gravitational waves are a cosmic phenomenon that are a result of the collision of highly dense objects. Study of gravitational waves has become possible with ultra sensitive instruments such as the Laser Interferometer Gravitational-Wave Observatory (LIGO)~\cite{aasi2015advanced} and Virgo~\cite{acernese2014advanced}.
These detectors can detect changes in length caused by gravitational waves less than the width of a proton~\cite{sigg2016advanced}. Naturally, these hyper sensitive instruments are prone to instrumental and environmental disturbances such as non Gaussian transients known as \textit{glitches} which can mimic gravitational waves. The frequency of occurrence of these glitches is so high that the chance of these glitches masking a gravitational wave is non-negligible~\cite{crowston2017gravity}. It is important to identify these glitches and eliminate them for proper gravitational wave detection. Project \textit{Gravity Spy}~\cite{zevin2017gravity} is an effort to identify and categorize these glitches into different classes based on the morphology of spectrograms with help from citizen scientists. While these glitches can be categorized into 22 classes currently, there is a possibility that new classes of glitches might get added in the future as the detectors undergo commissioning before each observation run~\cite{abbott2016characterization, abbott2017calibration}.

There have been previous attempts~\cite{shen2018glitch, colgan2020efficient, bahaadini2018machine} to classify glitches using supervised deep learning techniques, but in this work we take an unsupervised representation learning approach. Unsupervised representation learning can help alleviate the need for a large amount of labelled data and the need to identify new classes of glitches as they appear during the operation of LIGO. Disentangled representation learning, as a branch of unsupervised representation learning, has several advantages, as pointed out in \cite{kumar2017variational}: invariance, transferability, interpretability, and conditioning and intervention. A large portion of recent literature on disentanglement learning is based on Variational Autoencoders (VAEs). \citet{higgins2017beta} introduce $\beta$-VAEs, which penalizes the $KL$ divergence between the variational posterior and the prior using the hyperparameter $\beta$. Bottleneck-VAEs \cite{burgess2018understanding} increase the capacity of the information bottleneck as the training progresses thus offering better reconstructions than $\beta$-VAEs. In this work, we show that Bottleneck-VAEs and $\beta$-VAEs are closely connected and propose a decreasing schedule for the hyperparameter $\beta$ in $\beta$-VAEs that controls the information capacity similar to the hyperparameter $C$ in the objective function of Bottleneck-VAEs. In addition, we provide experimental evidence on \textit{Gravity Spy} dataset to show superior performance of our proposed VAE in unsupervised learning of glitches and advantages of using our proposed VAE when compared to Bottleneck-VAEs.


\begin{figure}[t]
\vskip 0.2in
\begin{center}
\includegraphics[width=0.28\columnwidth]{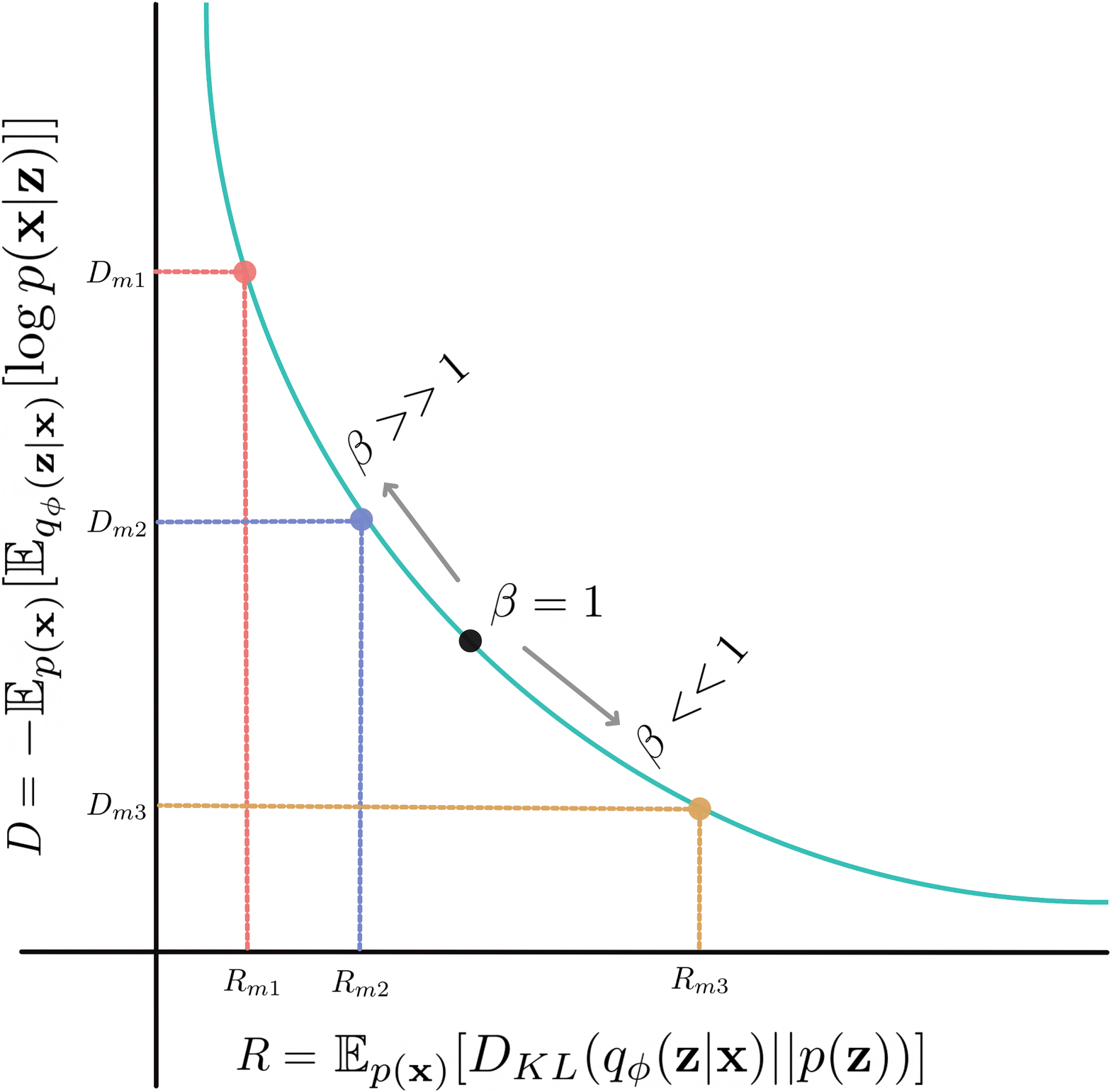}
\includegraphics[width=0.32\columnwidth]{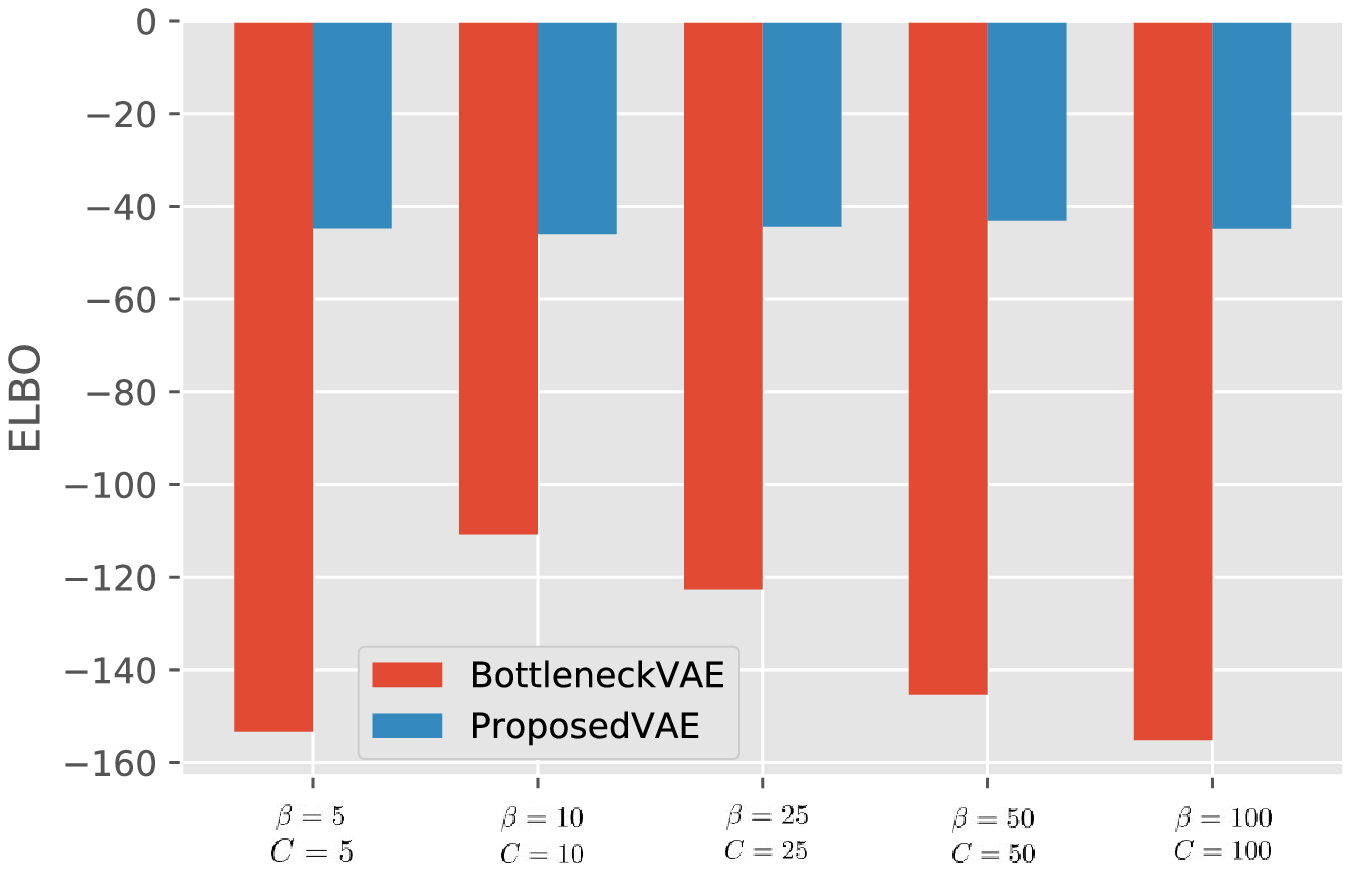}
\includegraphics[width=0.32\columnwidth]{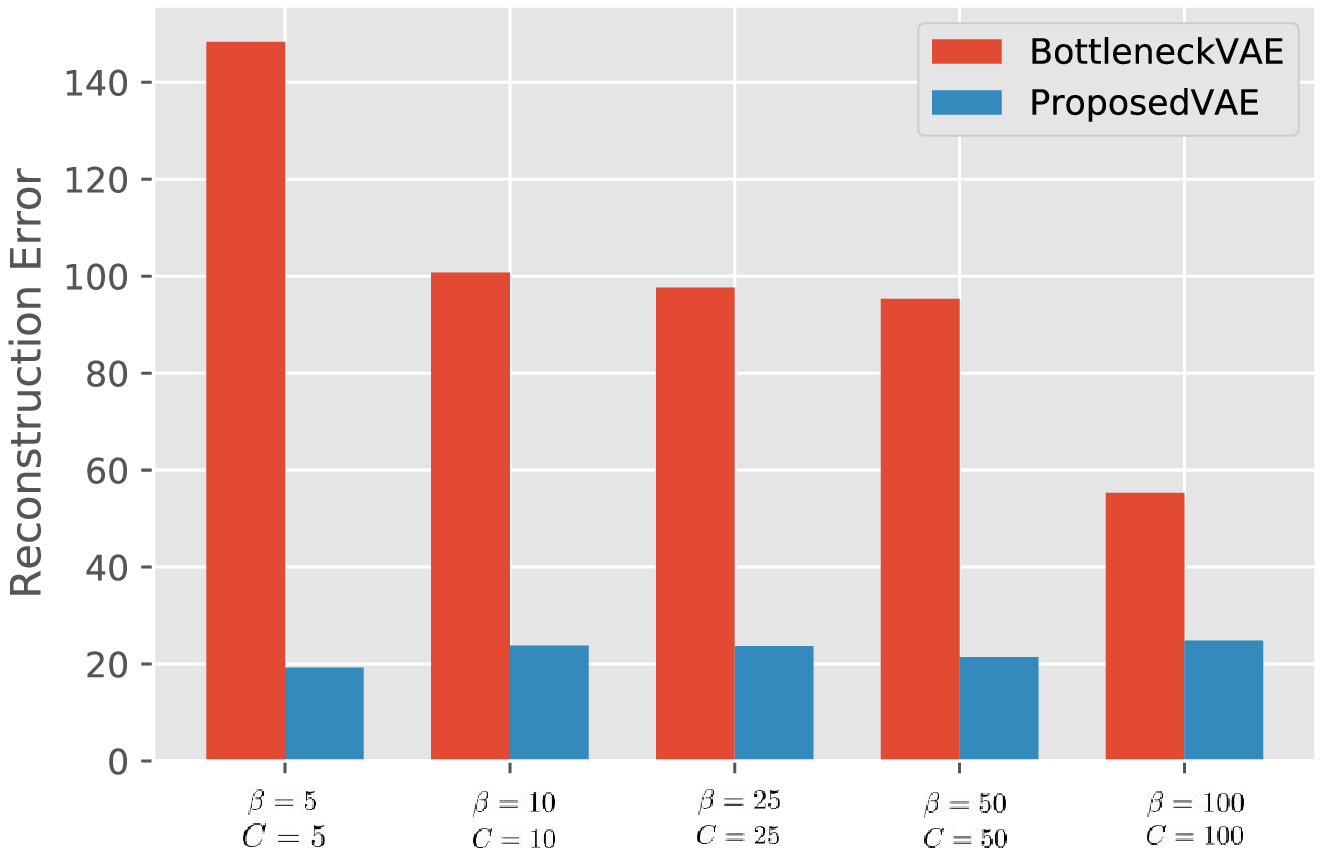}
\caption{(\textbf{Left}) Training Bottleneck VAEs with different values of $C$ equal to \textit{Rate} $R_{m_1}, R_{m_2}, R_{m_3}$ (with $\gamma=1$) corresponds to VAEs converging to points corresponding to \textit{Distortion} equal to $D_{m_1}, D_{m_2}, D_{m_3}$ in the $RD$-curve. ELBO (\textbf{Center}) and reconstruction error (\textbf{Right}) for different hyperparameters in Bottleneck and Proposed VAEs on dSpirtes.}
\label{fig:rd_curve}
\end{center}
\vskip -0.2in
\end{figure}

\section{VAE, $\beta$-VAE and Bottleneck-VAE}
The generative model of our data is defined as $p(\bm{x}|\bm{z})p(\bm{z}) = p(\bm{x},\bm{z})$ where each observed datapoint $\bm{x}$ is assumed to be generated from its own latent variable $\bm{z}$. VAEs
attempt to maximize the marginal likelihood of the data:
\begin{displaymath}
    \log p_\theta(\bm{x}_1,...,\bm{x}_N) = \sum_{i=1}^{N} \log p_\theta(\bm{x_i}). 
\end{displaymath}
Due to its intractability, a variational distribution $q_{\phi}$ is introduced to approximate the posterior $p(\bm{z}|\bm{x})$, which gives rise to the lower bound on the marginal likelihood called the \textit{Evidence Lower Bound} (ELBO):
\begin{equation}
    \mathcal{L}(\theta,\phi; \bm{x}, \bm{z}) = \mathbb{E}_{q_\phi(\bm{z}|\bm{x})}[\log p_\theta(\bm{x} | \bm{z})] - D_{KL}(q_\phi(\bm{z}|\bm{x})||p(\bm{z})).
\end{equation}
The integration in the first term is usually computed using samples from $q_\phi(\bm{z}|\bm{x})$ and backpropogation through the sampling process is done through the \textit{reparametrization trick}~\cite{kingma2013auto}. In practice, $q_\phi(\bm{z}|\bm{x})$ is assumed to be a Gaussian distribution with diagonal covariance and $p(\bm{z})$ to be $\mathcal{N}$(\textbf{0}, \textbf{I}).

$\beta$-VAEs \cite{higgins2017beta} are variants of regular VAEs that introduce a hyperparameter called $\beta$ to the ELBO:
\begin{equation}
    \mathbb{E}_{q_\phi(\bm{z}|\bm{x})}[\log p_\theta(\bm{x} | \bm{z})] - \beta D_{KL}(q_\phi(\bm{z}|\bm{x})||p(\bm{z})).
\end{equation}
For $\beta > 1$, $q_\phi(\bm{z}|\bm{x})$ is heavily constrained to be closer to the factorized prior $p(\bm{z})$. Heavy penalty on the $D_{KL}$ encourages
disentanglement, while at the same time leads to poor reconstruction quality. This is due to the fact that the latent factors are not able to encode enough information about the observations.

\citet{burgess2018understanding} proposed an alternate objective with an additional hyperparameter C; we call this variant of VAE as Bottleneck-VAE:
\begin{equation}
    \mathbb{E}_{q_\phi(\bm{z}|\bm{x})}[\log p_\theta(\bm{x} | \bm{z})] - \gamma | D_{KL}(q_\phi(\bm{z}|\bm{x})||p(\bm{z})) - C|.
    \label{eqn:bottleneck_vae}
\end{equation}
When $C=0$, the objective is the same as $\beta$-VAE, since $D_{KL} \ge 0$. In Bottleneck-VAEs, $C$ is progressively increased during training (with $\gamma$ kept constant, typically greater than one) to increase the amount of information stored about observations in the latent codes. This results in two effects: 1) As the information capacity is increased (through $C$) during training, the encoder learns to encode latent dimensions in the order of decreasing returns to log-likelihood. 2) This controlled capacity increase also encourages better reconstruction quality compared to $\beta$-VAEs, while achieving similar levels of disentanglement.


\section{$\beta$-Annealed VAE}
It is important to note that the objective functions corresponding to $\beta$-VAEs and Bottleneck-VAEs do not optimize the ELBO when $\beta > 1$ and $\gamma > 1, C > 0$, respectively. \cite{alemi2017fixing} offers an information theoretic perspective, in that $\beta$-VAEs 
try to find the optimal \textit{distortion} ($D$) and \textit{rate} ($R$) for a fixed $\beta = \frac{\partial D}{\partial R}$ by minimizing $\min_{q_\phi(\bm{z}|\bm{x}), p(\bm{z}), p_\theta(\bm{x} | \bm{z})} D + \beta R$, where $D = -\mathbb{E}_{p(\bm{x})}[\mathbb{E}_{q_\phi(\bm{z}|\bm{x})}[\log p_\theta(\bm{x} | \bm{z})]]$ and $R = \mathbb{E}_{p(\bm{x})}[D_{KL}(q_\phi(\bm{z}|\bm{x})||p(\bm{z}))]$. The inequality $ H - D \leq \mathcal{I}(X;Z) \leq R $ from \cite{alemi2017fixing} shows the relationship between $D$, $R$, \textit{data entropy} $H$ ($-\mathbb{E}_{p(\bm{x})}[\log p(\bm{x})]$) and \textit{mutual information} $\mathcal{I}$ ($D_{KL}( p(\bm{x, z}) || p(\bm{x}) p(\bm{z}) )$). For a finite capacity encoder $q_\phi(\bm{z}|\bm{x})$ and decoder $p_\theta(\bm{x} | \bm{z})$, vanilla VAEs correspond to 
an operating point on the green curve (in Figure.\ref{fig:rd_curve}) with slope 1. Fixing the capacity of the encoder and decoder, if $\beta$ is varied from greater than 1 to less than 1, the operating point shifts from $\uparrow D, \downarrow R$ to $\downarrow D, \uparrow R$ along the green curve (with $\uparrow$ \& $\downarrow$ denoting high and low respectively).



\begin{table}
\begin{minipage}{0.69\linewidth}
\centering
\captionsetup{type=figure}
	\includegraphics[width=0.49\columnwidth]{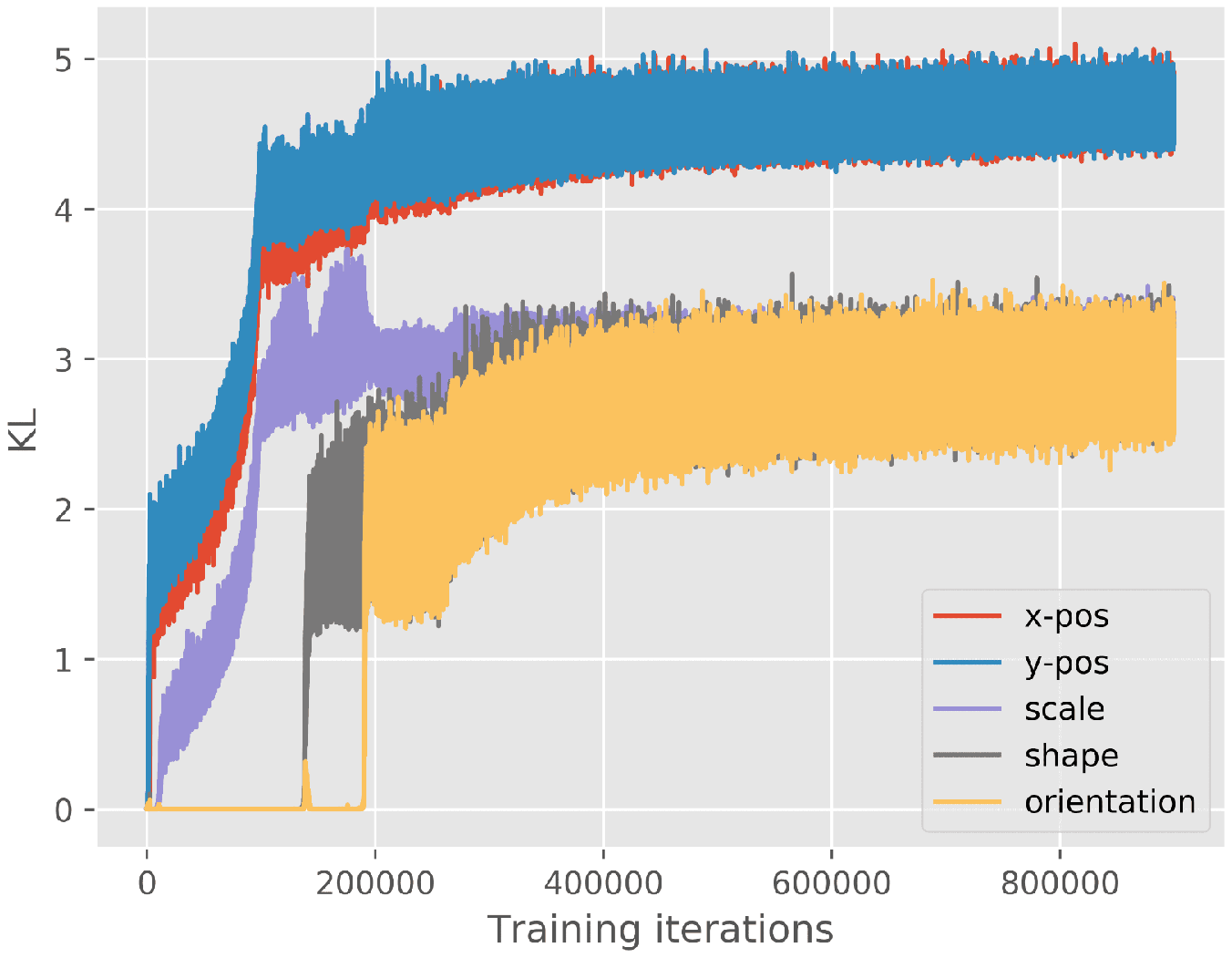}
    \includegraphics[width=0.49\columnwidth]{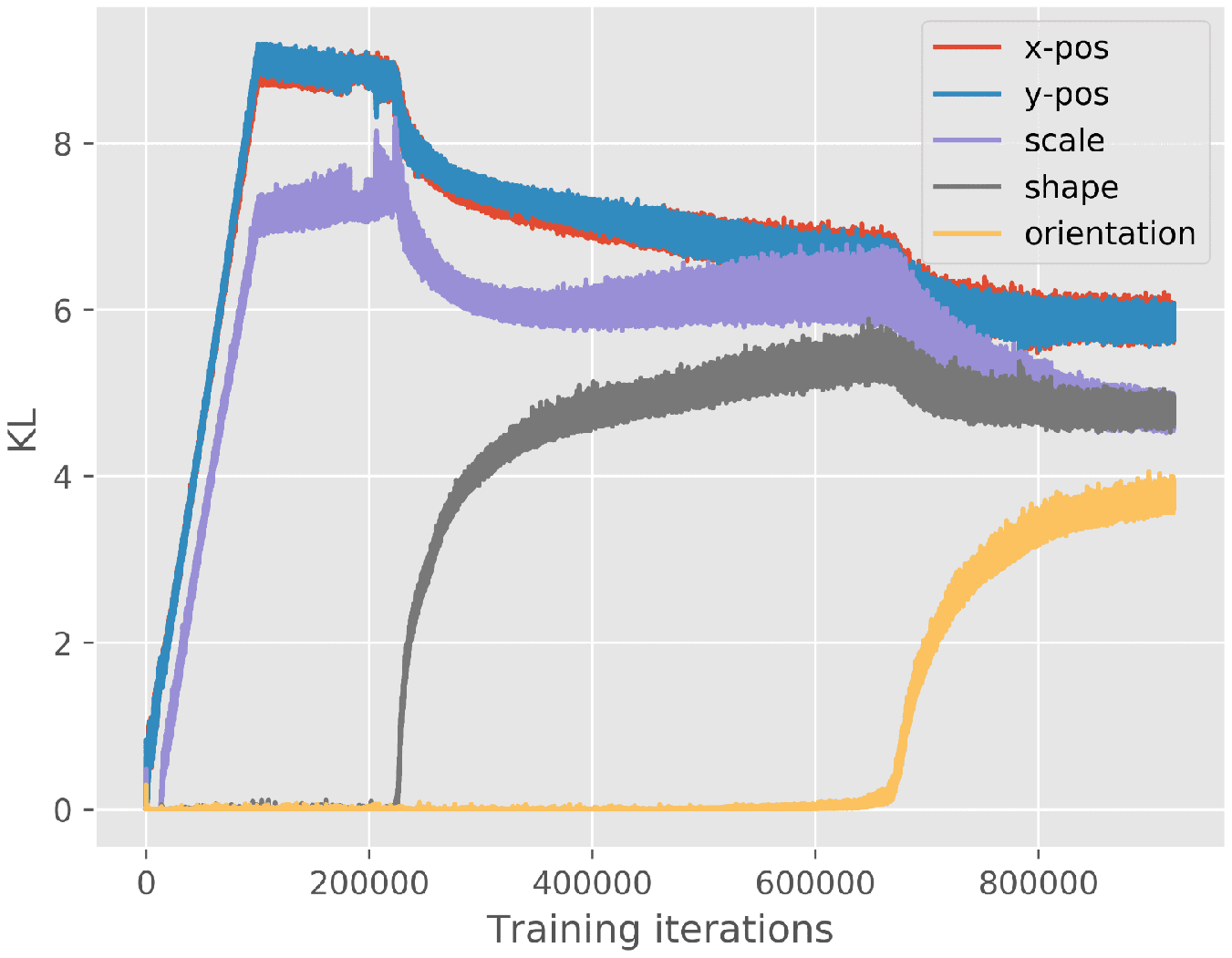}
    \caption{$KL$-divergence of each latent dimension with respect to a unit Gaussian during training on dSprites. \textbf{Left:} In $\beta$-Annealed VAE, $\beta$ is decreased as the training progresses \textbf{Right:} In Bottleneck-VAE, $C$ is increased as the training progresses to increase the information capacity}
	\label{fig:one_latent_dim_at_a_time}
\end{minipage}\hfill
\begin{minipage}{0.29\linewidth}
	\caption{Results on unsupervised representation learning of non Gaussian noise transients that occur in gravitational wave detectors}
    \vspace{0.5cm}
	\label{table:student}
	\centering
	\resizebox{0.87\linewidth}{!}{%
	\begin{tabular}{cc}
      \hline
      Model & Accuracy \\ \hline
        $\beta$-VAE & 61.26\% \\
        Bottleneck-VAE & 80.01\% \\
        Proposed-VAE & \textbf{81.60}\% \\ \hline
      \end{tabular}%
      }
     \label{table:ligo_results}
\end{minipage}
\end{table}

We first view Bottleneck-VAE from an information theoretic standpoint. If we set $\gamma = 1$ and a constant $C = R_m$, optimizing (\ref{eqn:bottleneck_vae}) can be viewed as minimizing $D$ for a constant $R = R_m$. From Figure~\ref{fig:rd_curve}, we can see
that this corresponds to a point on the $RD$ curve where $R = R_m$. For different increasing values of $C$, the point shifts to locations in the $RD$ curve corresponding to increasing $R$. Concretely, increasing $C$ corresponds to relaxing the constraint that $q_\phi(\bm{z}|\bm{x})$ needs to be closer in terms of $KL$-divergence to the prior $p(\bm{z}))$, and \cite{burgess2018understanding} showed that when $\gamma > 1$, this leads to better robust disentanglement and better reconstruction quality.

Since any point on the $RD$-curve for a fixed encoder and decoder is reachable through $\beta$, controllable information capacity can be achieved through $\beta$. Motivated by the monotonically increasing schedule of $C$ in case of Bottleneck-VAEs, we propose a monotonically decreasing schedule of $\beta$ for $\beta$-VAEs. The effect on distortion and rate while decreasing $\beta$ in case of $\beta$-VAEs is very similar to the effect of increasing $C$ in Bottleneck-VAEs. We formalize our claim in the following lemma (Proof in Appendix),

\begin{lemma}
(For a fixed finite capacity encoder and decoder) Let $D_{C_1}^*, D_{C_2}^*$ and $R_{C_1}^*, R_{C_2}^*$ denote the optimal distortion and rate for a Bottleneck-VAE with $C_1, C_2$ respectively with a constant $\gamma \geq 0$. Similarly let $D_{\beta_1}^*, D_{\beta_2}^*$ and $R_{\beta_1}^*, R_{\beta_2}^*$ denote the optimal distortion and rate for a $\beta$-VAE with $\beta_1, \beta_2$ respectively. If $C_1 > C_2 \geq 0$, then $R_{C_1}^* > R_{C_2}^*$ and $D_{C_1}^* < D_{C_2}^*$. Similarly, with respect to $\beta$-VAEs, if  $0 \leq \beta_1 < \beta_2$, then $R_{\beta_1}^* > R_{\beta_2}^*$ and $D_{\beta_1}^* < D_{\beta_2}^*$. 
\end{lemma}

If we want to replicate similar effects of linearly increasing $C$ in the case of Bottleneck-VAEs, a $\beta$-VAE can be trained with monotonically decreasing $\beta$ from $\beta \gg 1$ to $\beta \ll 1$. We use linearly decreasing schedule for $\beta$ in all of our experiments. When compared to Bottleneck-VAEs, a linearly decreasing schedule of $\beta$ in $\beta$-VAEs (which we call $\beta$-Annealed VAEs) offers advantages such as: 1) without having to set C, our proposed schedule have one less hyperparameter to tune; 2) in all of our experiments, we linearly decreasing $\beta$ from $\beta \gg 1$ to $\beta = 1$ during training, which can be interpreted as $\beta$-VAEs are trained as vanilla VAEs during later stages in training leading to better reconstruction error.

\begin{table*}[!]
\caption{Quantitative assessment of disentanglement in dSprites}
\label{sample-table}
\vskip 0.15in
\begin{center}
\begin{small}
\begin{sc}
\resizebox{0.85\linewidth}{!}{%
\begin{tabular}{lcccccccr}
\toprule
VAE & Hyperparameter & BetaVAE & FactorVAE & MIG & DCI & Modularity & SAP \\
Variant & & Score & Score & & Disentanglement & & \\
\midrule
\multirow{3}{*}{$\beta$-VAE \cite{higgins2017beta}} & $\beta$=1 & 0.851 & 0.685 & 0.072 & 0.127 & 0.790 & 0.052\\
&$\beta$=4 & 0.816 & 0.627 & 0.078 & 0.138 & 0.800 & 0.028 \\ 
& $\beta$=16 & 0.742 & 0.546 & 0.141 & 0.277 & 0.809 & 0.010 \\ \\
\multirow{3}{*}{Bottleneck-VAE \cite{burgess2018understanding}} & $C$=5 & 0.868 & 0.596 & \textbf{0.334} & 0.402 & 0.791 & \textbf{0.078}\\
&$C$=25 & 0.765 & 0.539 & 0.025 & 0.059 & 0.769 & 0.022 \\
& $C$=100 & 0.625 & 0.369 & 0.014 & 0.022 & 0.746 & 0.007 \\ \\
\multirow{3}{*}{Factor-VAE \cite{kim2018disentangling}} & $\gamma$=10 & 0.862 & 0.706 & 0.144 & 0.221 & 0.781 & 0.068 \\
& $\gamma$=30 & 0.878 & \textbf{0.849} & 0.190 & 0.328 & 0.796 & 0.068 \\
& $\gamma$=100 & 0.862 & 0.792 & 0.312 & \textbf{0.461} & 0.820 & 0.062 \\ \\
\multirow{3}{*}{$\beta$-TCVAE \cite{chen2018isolating}} & $\beta$=1 & 0.851 & 0.685 & 0.072 & 0.127 & 0.790 & 0.052 \\
&$\beta$=4 & 0.875 & 0.830 & 0.226 & 0.347 & 0.805 & 0.064 \\ 
& $\beta$=10 & 0.879 & 0.808 & 0.287 & 0.447 & 0.818 & 0.067 \\ \\
\multirow{3}{*}{DIP-VAE-I \cite{kumar2017variational}} & $\lambda_{od}$=1 & 0.846 & 0.645 & 0.094 & 0.127 & 0.779 & 0.053 \\ &$\lambda_{od}$=5 & 0.804 & 0.574 & 0.040 & 0.077 & 0.783 & 0.025 \\ 
& $\lambda_{od}$=50 & 0.783 & 0.599 & 0.034 & 0.077 & 0.778 & 0.016 \\ \\
\multirow{3}{*}{DIP-VAE-II \cite{kumar2017variational}} & $\lambda_{od}$=1 & 0.720 & 0.479 & 0.015 & 0.083 & 0.782 & 0.004\\ &$\lambda_{od}$=5 & 0.793 & 0.644 & 0.049 & 0.108 & 0.798 & 0.016\\ 
& $\lambda_{od}$=50 & 0.869 & 0.544 & 0.087 & 0.177 & 0.809 & 0.058\\ \\
\multirow{3}{*}{Proposed VAE} & $\beta$=5 & 0.846 & 0.809 & 0.073 & 0.180 & 0.815 & 0.038\\ 
&$\beta$=25 & 0.739 & 0.599 & 0.111 & 0.233 & 0.790 & 0.034\\
& $\beta$=50 & \textbf{0.902} & 0.805 & 0.289 & 0.397 & \textbf{0.832} & 0.076\\ \\
\bottomrule
\end{tabular}%
}
\end{sc}
\end{small}
\end{center}
\label{table:disentanglement_results}
\vskip -0.1in
\end{table*}


\section{Results}
We perform experiments to indicate $\beta$-Annealed VAEs behave similarly to Bottleneck-VAEs when the information capacity is increased. Then we compare $\beta$-Annealed VAEs and Bottleneck-VAEs in terms of ELBO, reconstruction error and disentanglement on the dSprites~\cite{dsprites17} (qualitative assessment of disentanglement can be found in Appendix).

We first show that the two effects of linearly increasing C in Bottleneck-VAEs (with $\gamma > 1$ kept constant) can be achieved using linearly decreasing $\beta$ in $\beta$-VAEs. We use the same architecture of encoder and decoder used in \cite{burgess2018understanding} and trained a $\beta$-VAE with linearly decreasing $\beta$ from 100 to 1 (with iteration threshold being 100000). Figure~\ref{fig:one_latent_dim_at_a_time} shows the $D_{KL}$ of each latent dimension $q(z|\bm{x})$ to its prior (standard normal distribution), we see that the generative factors are learned one at a time by the network in the order of decreasing returns to the log-likelihood, similar to Bottleneck-VAEs. To show that $\beta$-Annealed VAEs  achieve better reconstruction error because they are trained as vanilla VAEs during the later stages of training (i.e after $\beta$ is reduced to 1), we perform experiments with different values of $\beta$ and $C$, and Figure~\ref{fig:rd_curve} (\textit{right}) shows the reconstruction error on dSpirtes for Bottleneck-VAEs and $\beta$-VAEs. We see that our proposed linearly decreasing schedule of $\beta$ offers better reconstruction error than Bottleneck-VAEs. Figure~\ref{fig:rd_curve} (\textit{center}) also shows that $\beta$-Annealed VAEs achieve better ELBO than Bottleneck-VAEs. To quantitatively assess disentanglement offered by our proposed linear decreasing schedule of $\beta$ in $\beta$-VAEs, we used the metrics \textit{$\beta$-VAE metric} \cite{higgins2017beta}, \textit{Factor VAE metric} \cite{kim2018disentangling}, \textit{Mutual Information Gap (MIG)} \cite{chen2018isolating}, \textit{Modularity} \cite{ridgeway2018learning}, \textit{DCI Disentanglement} \cite{eastwood2018framework} and \textit{SAP score} \cite{kumar2017variational}, similar to \cite{locatello2018challenging}. We show the disentanglement performance of the proposed method with the following existing variants of VAEs: 1) $\beta$-VAE, 2) FactorVAE, 3) TCVAE, 4)DIP-VAE-I, 5)DIP-VAE-II, 6) Bottleneck-VAE in Table \ref{table:disentanglement_results}.

Further, we train $\beta$-VAE, Bottleneck-VAE and $\beta$-Annealed VAE on the \textit{Gravity Spy} \cite{zevin2016gravity} dataset, which contains spectrogram samples of 22 different types of glitches. We check the quality of representations learnt by the encoders by training linear classifiers trained on top of latent representations and their performance are as shown in Table \ref{table:ligo_results}. We see that our proposed VAE learns better representations when compared to $\beta$-VAEs and Bottleneck-VAEs.


\section{Conclusion}
We introduce $\beta$-Annealed VAEs motivated by viewing Bottlenck-VAEs through the lens of information theory. We show that our proposed version of $\beta$-VAEs, with linearly decreasing $\beta$ as the training progresses, offers similar robust disentanglement while having better reconstruction error.
We prove its efficacy in learning representations of glitches in LIGO / Virgo detectors.

\section{Broader Impact}

Beyond unsupervised representation learning of glitches in gravitational wave detectors, $\beta$-Annealed VAEs can be used in applications requiring disentanglement of generative factors of data. Since our proposed VAEs have lower reconstruction errors, they can be used in applications where sample quality is important. We believe this work could encourage the ML community to delve into unsupervised learning techniques for the detection and study of glitches. We see research opportunities in devising specific types of VAEs after closely studying the characteristics of glitches and developing a standard benchmark to test different models. 

\bibliography{neurips_2020}
\bibliographystyle{unsrtnat}

\newpage
\appendix
\section{Qualitative assessment}

\begin{figure}[h]
\begin{center}
\centerline{\includegraphics[width=0.65\columnwidth]{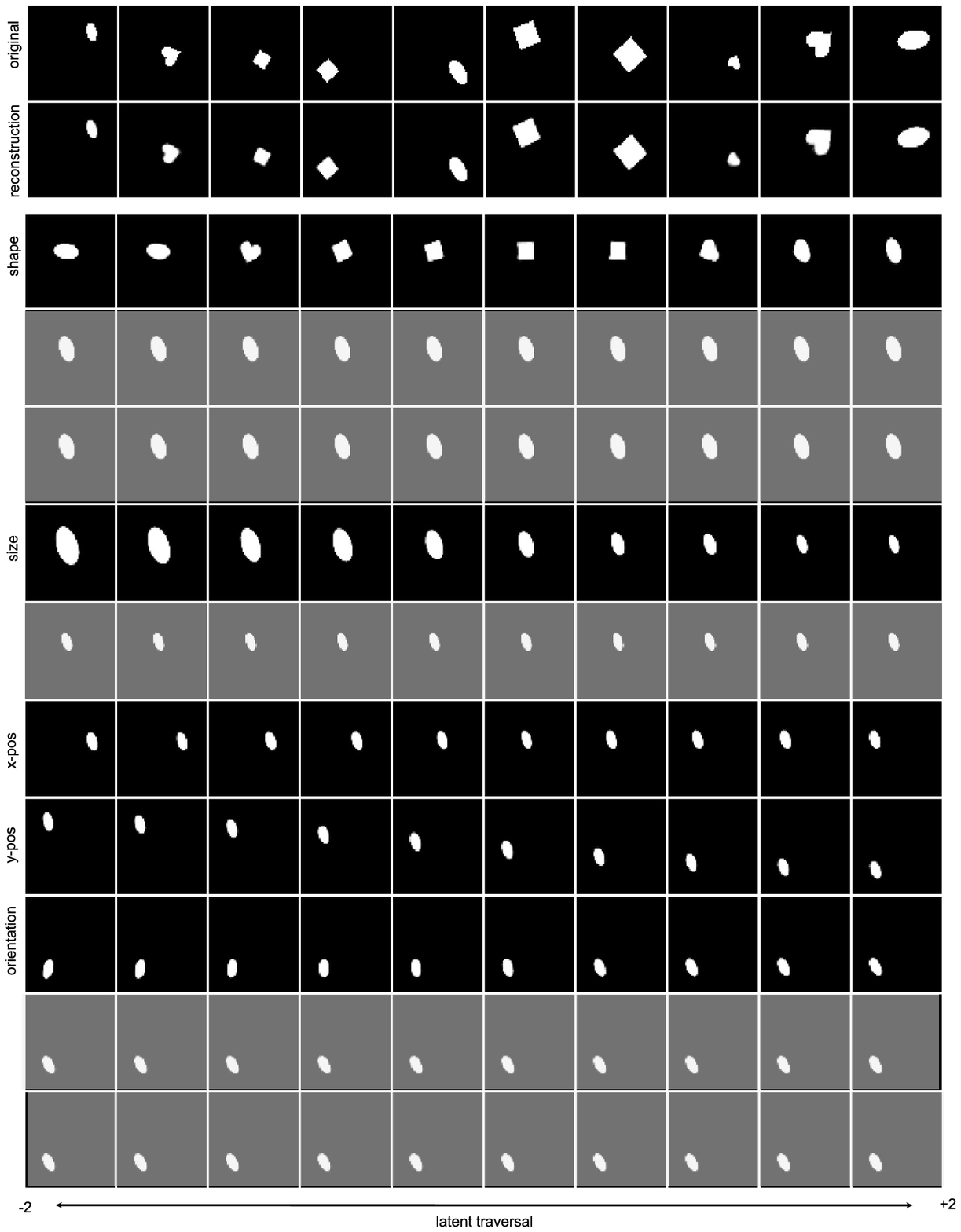}}
\caption{\textbf{First two rows:} Original data and their reconstructions \textbf{Other rows:} all 10 latent dimension traversals with captured attribute indicated in the sides. Greyed out rows indicate dead dimensions.}
\label{fig:dspirtes_latent_traversals}
\end{center}
\vskip -0.3in
\end{figure}

\begin{figure*}[b]
\begin{center}
\centerline{\includegraphics[width=\linewidth]{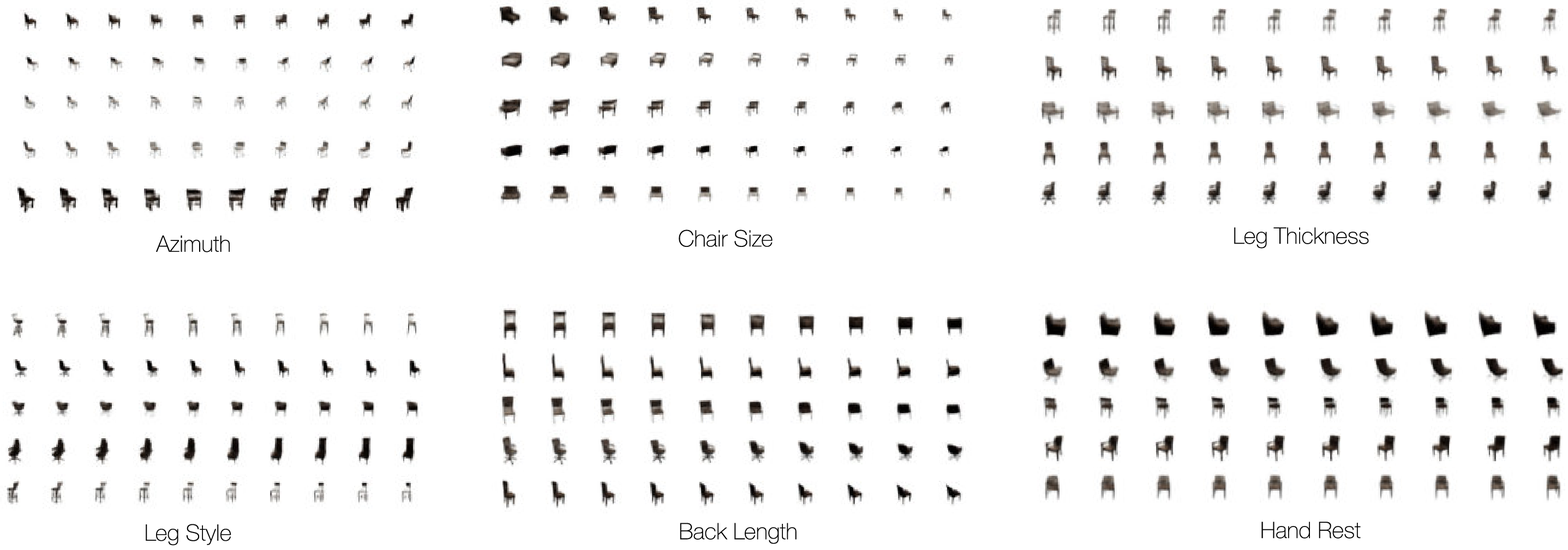}}
\caption{Latent traversals of different latent dimensions on 3DChairs dataset with traversals in the range [-2, 2] using $\beta$-Annealed VAEs with $\beta=50$}
\label{fig:3dchairs_traversals}
\end{center}
\end{figure*}

\newpage
\section{Proofs}

\begin{lemma}
(For a fixed finite capacity encoder and decoder) Let $D_{C_1}^*, D_{C_2}^*$ and $R_{C_1}^*, R_{C_2}^*$ denote the optimal distortion and rate for a Bottleneck-VAE with $C_1, C_2$ respectively with a constant $\gamma \geq 0$. Similarly let $D_{\beta_1}^*, D_{\beta_2}^*$ and $R_{\beta_1}^*, R_{\beta_2}^*$ denote the optimal distortion and rate for a $\beta$-VAE with $\beta_1, \beta_2$ respectively. If $C_1 > C_2 \geq 0$, then $R_{C_1}^* > R_{C_2}^*$ and $D_{C_1}^* < D_{C_2}^*$, similarly with respect to $\beta$-VAEs, if  $0 \leq \beta_1 < \beta_2$, then $R_{\beta_1}^* > R_{\beta_2}^*$ and $D_{\beta_1}^* < D_{\beta_2}^*$ 
\end{lemma}

\textit{Proof.} From the objective function of Bottleneck-VAEs,
\begin{displaymath}
    \min_{q_\phi(\bm{z}|\bm{x}), p(\bm{z}), p(\bm{x} | \bm{z})} D + \gamma|R - C|
\end{displaymath}

one can see that the optimal values for R, when $C = C_1$ is $R_{C_1}^* = C_1$ (similarly, when $C = C_2$, $R_{C_2}^* = C_2$). If $C_1 > C_2$ then $R_{C_1}^* > R_{C_2}^*$. Also,

\begin{align*}
    H - D_{C_2}^* &\leq R_{C_2}^* < R_{C_1}^* \\
    H - D_{C_2}^* &< H - D_{C_1}^* \\
    D_{C_1}^* &< D_{C_2}^*
\end{align*}

For $\beta$-VAEs, if $\beta_1 < \beta_2$, then $R_{\beta_1}^* > R_{\beta_2}^*$ and $D_{\beta_1}^* < D_{\beta_2}^*$ is a direct result from \cite{alemi2017fixing}. $\beta$-VAEs with a fixed architecture and finite capacity can be used to interpolate between auto-encoding behaviour ($\uparrow D, \downarrow R$) to auto-decoding ($\downarrow D, \uparrow R$) behaviour by changing from $\beta << 1$ to $\beta >> 1$.










\end{document}